\documentclass{article}

\usepackage{microtype}
\usepackage{graphicx}
\usepackage{subcaption}
\usepackage{booktabs} %
\usepackage{multirow}
\usepackage{makecell}

\usepackage{hyperref}

\usepackage[accepted]{icml2024}

\usepackage{amsmath}
\usepackage{amssymb}
\usepackage{mathtools}
\usepackage{amsthm}

\usepackage[capitalize,noabbrev]{cleveref}

\usepackage{amsmath,amsfonts,bm}

\def\1{\bm{1}}

\def\vzero{{\bm{0}}}

\def\vtheta{{\bm{\theta}}}

\def\veps{{\bm{\epsilon}}}

\def\vm{{\bm{m}}}

\def\vs{{\bm{s}}}

\def\vx{{\bm{x}}}
\def\vy{{\bm{y}}}
\def\vz{{\bm{z}}}

\def\mX{{\bm{X}}}

\def\mZ{{\bm{Z}}}

\DeclareMathAlphabet{\mathsfit}{\encodingdefault}{\sfdefault}{m}{sl}
\SetMathAlphabet{\mathsfit}{bold}{\encodingdefault}{\sfdefault}{bx}{n}

\def\hvy{{\hat{\bm{y}}}}
\def\hmY{{\hat{\bm{Y}}}}

\newcommand{\R}{\mathbb{R}}

\newcommand\abs[1]{\left\lvert#1\right\rvert}

\DeclarePairedDelimiterX{\infdivx}[2]{(}{)}{%
  #1\;\delimsize\|\;#2%
}

\DeclarePairedDelimiter{\norm}{\lVert}{\rVert}

\DeclareMathOperator*{\argmax}{arg\,max}

\DeclareMathOperator*{\E}{\mathbb{E}}

\DeclareMathOperator{\onehot}{one\,hot}

\theoremstyle{plain}

\theoremstyle{definition}

\theoremstyle{remark}

\usepackage[most,breakable]{tcolorbox}
\usepackage{etoolbox}
\AtBeginEnvironment{tcolorbox}{\small}

\newcommand{\smallsc}[2][]{\textsc{\small #2}}

\icmltitlerunning{Forget Sharpness: Perturbed Forgetting of Model Biases Within SAM Dynamics}

\begin{document}

\twocolumn[
\icmltitle{Forget Sharpness: Perturbed Forgetting of Model Biases Within SAM Dynamics}

\icmlsetsymbol{intern}{*}

\begin{icmlauthorlist}
\icmlauthor{Ankit Vani}{mila,bai,intern}
\icmlauthor{Frederick Tung}{bai}
\icmlauthor{Gabriel L. Oliveira}{bai}
\icmlauthor{Hossein Sharifi-Noghabi}{bai}
\end{icmlauthorlist}

\icmlaffiliation{mila}{Mila, Universit\'e de Montr\'eal}
\icmlaffiliation{bai}{Borealis AI}

\icmlcorrespondingauthor{Ankit Vani}{ankit.vani@umontreal.ca}

\icmlkeywords{Machine Learning, ICML}

\vskip 0.3in
]

\printAffiliationsAndNotice{\textsuperscript{*}Work done during an internship at Borealis AI }

\begin{abstract}

Despite attaining high empirical generalization, the sharpness of models trained with sharpness-aware minimization (SAM) do not always correlate with generalization error. Instead of viewing SAM as minimizing sharpness to improve generalization, our paper considers a new perspective based on SAM's training dynamics. We propose that perturbations in SAM perform \emph{perturbed forgetting}, where they discard undesirable model biases to exhibit learning signals that generalize better. We relate our notion of forgetting to the information bottleneck principle, use it to explain observations like the better generalization of smaller perturbation batches, and show that perturbed forgetting can exhibit a stronger correlation with generalization than flatness. While standard SAM targets model biases exposed by the steepest ascent directions, we propose a new perturbation that targets biases exposed through the model's outputs. Our output bias forgetting perturbations outperform standard SAM, GSAM, and ASAM on ImageNet, robustness benchmarks, and transfer to CIFAR-\{10,100\}, while sometimes converging to sharper regions. Our results suggest that the benefits of SAM can be explained by alternative mechanistic principles that do not require flatness of the loss surface.

\end{abstract}

\section{Introduction}
\label{sec:intro}

\begin{figure}[t]
\vskip 0.2in
\begin{center}
\centerline{
\includegraphics[width=0.95\columnwidth]{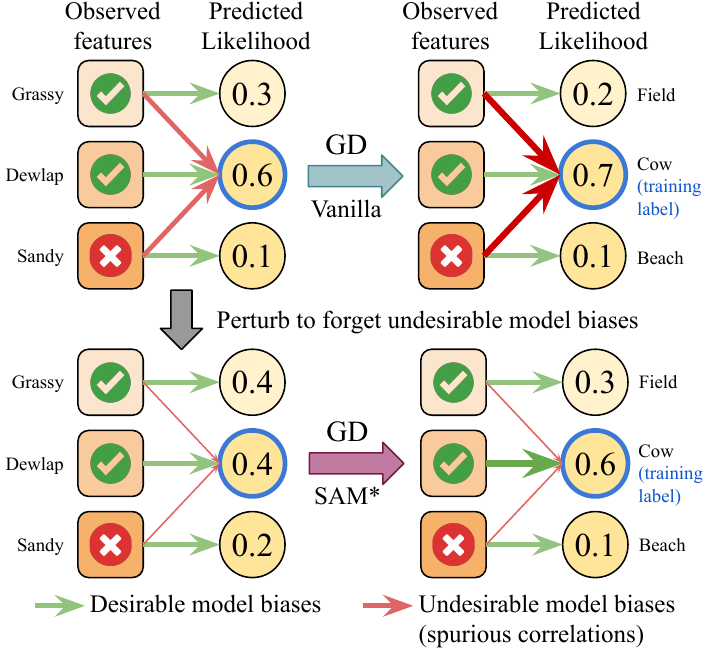}
}
\caption{A simplified illustration of our mechanistic \emph{perturbed forgetting} perspective of sharpness-aware minimization (SAM). We treat perturbations in each step of SAM as an opportunity to forget undesirable model biases. Here, the presence of `grassy' or `sandy' features spuriously contribute to the prediction of `cow.' When gradient descent (GD) can strengthen these biases, leading to overfitting, the perturbation of SAM takes an ascent step to `forget' them to allow computing a less biased gradient. *Not illustrated: this gradient is used to take a GD step at the \emph{unperturbed} weights.}
\label{fig:samoverview}
\end{center}
\vskip -0.2in
\end{figure}

The belief that flatter minima of the loss surface generalize better is commonplace in machine learning \cite{jiang2019fantastic}. Sharpness-aware minimization (SAM) \cite{foret2020sharpness} and its variants \cite{kwon2021asam,zhuang2022surrogate,kim2022fisher} are motivated and presented as methods to minimize sharpness to improve generalization. As many models trained with SAM exhibit better generalization, research continues to explore the principles behind it \cite{andriushchenko2022towards,wen2022sharpness} to improve training algorithms. However, certain questions stand in the way of refining these techniques. First, the sharpness metric induced by SAM does not necessarily correlate with generalization in modern deep learning architectures \cite{andriushchenko2023modern,kaur2023maximum}. Furthermore, practically necessary concepts like $m$-sharpness are unsupported by the theory these methods are based on, casting doubt on the potential of progress upon assumptions that do not hold up empirically.

Instead of looking at these methods from the perspective of reducing sharpness, we offer a novel view by considering a mechanistic aspect of SAM's training dynamics, which we term \emph{perturbed forgetting}. Each update step in SAM comprises of perturbing the model parameters with a gradient ascent step, and using the gradients computed at these perturbed parameters to update the original weights. Under our perspective, we treat the perturbations as an opportunity to `forget' undesirable model biases, as illustrated in \cref{fig:samoverview}. While such model biases are forgotten only during perturbation, we can reinterpret concepts like minimizing the surrogate gap using GSAM \cite{zhuang2022surrogate} to provide explicit mechanisms to unlearn unnecessary biases. Our perspective also offers explanations for the generalization benefits of small perturbation batches (low $m$ in $m$-sharpness) and the importance of worst-case perturbations over random ones with SAM \cite{andriushchenko2022towards}, which we discuss in more detail in \cref{sec:perspective}.

The biases a model learns can be exposed by probing its various aspects like gradients and outputs. Under the perturbed forgetting perspective, standard SAM perturbations target model biases exposed in the steepest gradient directions of small batches for forgetting. However, when we consider that a model's outputs can also expose model biases, we argue that the steepest ascent perturbation can have the opposite effect and amplify them. To address this limitation, we propose an output bias forgetting (OBF) perturbation in \cref{sec:method} that avoids amplifying these output-exposed biases, and optionally allows stronger forgetting by pushing predictions towards a uniform distribution. The success of our proposal suggests that non-standard probing mechanisms can be devised to target model biases in settings where the benefits of SAM are absent or minimal.

The notion of forgetting is related to the information bottleneck principle \cite{tishby2000information}, which suggests that optimal generalization may occur when a model retains only the information relevant to the task. Accordingly, the amount of task-irrelevant information discarded during perturbation enables us to quantify perturbed forgetting. We justify perturbed forgetting as an alternative to the narrative of sharpness minimization by showing that this quantity correlates with generalization more strongly than loss surface flatness in \cref{sec:forgetvsacc}. Other forgetting techniques have been proposed in the literature \cite{zhou2022fortuitous,ash2020warm,taha2021knowledge,tiwari2023overcoming} which modify parameters in-place, but the dynamics of SAM offer the advantage of transient forgetting for computing updates without disrupting the learning state.

We summarize our contributions in this paper as follows\footnote{Source code: \url{https://github.com/BorealisAI/perturbed-forgetting}.}:
\begin{enumerate}
    \item We present the perturbed forgetting perspective of SAM. We relate perturbed forgetting to generalization based on the information bottleneck principle, argue how standard SAM perturbations decrease an information-theoretic generalization bound, and empirically validate that forgetting can correlate with generalization better than loss surface flatness.
    \item Embracing the perturbed forgetting perspective, we design the OBF perturbation that targets model biases exposed in the model's outputs. Despite not necessarily exhibiting the lowest sharpness, our perturbation leads to improved generalization with the SAM, GSAM, and ASAM frameworks on ImageNet \cite{deng2009imagenet} and robustness benchmarks using ViTs \cite{dosovitskiy2020image} and ResNets \cite{he2016deep}.
\end{enumerate}
Our results suggest that the training dynamics of SAM may be more important than minimizing loss surface sharpness. The pursuit of flat minima could be a red herring, and the benefits of SAM's training dynamics might be better explained by other mechanistic principles.

\section{Background}
\label{sec:background}

We start by briefly detailing the preliminary concepts that we refer to in this paper.

\subsection{Sharpness-Aware Minimization (SAM)}

Sharpness-aware minimization (SAM) \cite{foret2020sharpness} is an optimization procedure that aims to minimize a PAC-Bayes upper-bound of the generalization error by considering perturbations of the model parameters. Let us represent a batch of $n$ samples drawn from the data distribution $\mathcal{D}$ as $S \sim \mathcal{D}^n$. Then, for a loss function $L_\vtheta(S)$ parameterized by $\vtheta \in \R^d$, the sharpness-aware optimization problem is
\begin{align}
&\min_\vtheta L_\vtheta^{\text{SAM}}(S) + \lambda \norm{\vtheta}_2^2,\label{eq:samopt}\\
\text{where }\, &L_\vtheta^{\text{SAM}}(S) = \max_{\norm{\veps}_2 \leq \rho} L_{\vtheta + \veps}(S).\label{eq:samperturb}
\end{align}
Here, $\lambda$ is an L2-regularization hyperparameter, and $\rho$ is a hyperparameter that controls for the neighborhood size for the perturbation.

To make this min-max problem tractable with stochastic gradient descent, SAM approximates the inner maximization problem by considering a first-order Taylor approximation of $L_{\vtheta + \veps}(S)$ w.r.t. $\veps$ around $\vzero$, giving us the gradient
\begin{align}
\nabla_\vtheta L_\vtheta^{\text{SAM}}(S) \approx \nabla_\vtheta L_\vtheta(S) \left|_{\vtheta + \rho \frac{\nabla_\vtheta L_\vtheta(S)}{\norm{\nabla_\vtheta L_\vtheta(S)}_2}} \right..\label{eq:samgrad}
\end{align}

A variant of SAM called $m$-SAM considers multiple perturbations using $m$-sized subsets of a training batch, which generalizes better than SAM in practice when $m$ is small \cite{foret2020sharpness,andriushchenko2022towards,wen2022sharpness}. $m$-SAM exhibits the update gradient:

\begin{align}
\nabla_\vtheta L_\vtheta^{\text{$m$-SAM}}(S) &= \E_{\tilde{S}\sim S^m}\left[ \nabla_\vtheta L_\vtheta(S) \left|_{\vtheta + \rho \frac{\nabla_\vtheta L_\vtheta(\tilde{S})}{\norm{\nabla_\vtheta L_\vtheta(\tilde{S})}_2}} \right.\right],\label{eq:msamgrad}
\end{align}
and the associated sharpness metric is termed $m$-sharpness \cite{foret2020sharpness}.

\subsection{Surrogate Gap Minimization with SAM (GSAM)}

GSAM \cite{zhuang2022surrogate} defines the surrogate gap $h(\vtheta)$ as the difference between the maximum loss within an $\epsilon$-neighborhood of parameters $\vtheta$ and the loss at $\vtheta$:
\begin{align}
    h(\vtheta) &= \max_{\norm{\veps}_2 \leq \rho} L_{\vtheta + \veps}(S) - L_\vtheta(S).\label{eq:sgap}
\end{align}
The authors show that the loss surface is flatter as $h$ gets closer to zero. To minimize the surrogate gap, the gradient $\nabla_\vtheta L_\vtheta(S)$ is first decomposed into components parallel and orthogonal to $\nabla_\vtheta L_\vtheta^{\text{SAM}}(S)$. Denoting the orthogonal component as $\nabla_\vtheta L_\vtheta^{\text{gap}}(S)$, the GSAM update gradient is
\begin{align}
\nabla_\vtheta L_\vtheta^{\text{GSAM}}(S) = \nabla_\vtheta L_\vtheta^{\text{SAM}}(S)-\xi \nabla_\vtheta L_\vtheta^{\text{gap}}(S),\label{eq:gsamgrad}
\end{align}
where $\xi$ is a hyperparameter that controls the step size in the direction of closing the surrogate gap.

\subsection{Information Bottleneck in Deep Learning}

The information bottleneck principle \cite{tishby2000information} describes the minimal sufficient statistics of an input random variable $\mX \in \mathcal{X}$ w.r.t. a target random variable $Y \in \mathcal{Y}$. In a neural network with $L$ layers, it suggests that an optimal representation $\mZ_l \in \mathcal{Z}_l$ for any $l \in [L]$ minimizes $I(\mX;\mZ_l) - \beta I(\mZ_l;Y)$, where $I$ denotes mutual information and $\beta$ trades off the representation complexity $I(\mX;\mZ_l)$ with the amount of relevant target information $I(\mZ_l;Y)$.

In their work justifying the benefit of information bottleneck in deep learning, \citet{kawaguchi2023does} present information theoretic bounds for generalization errors in neural networks comprised of $L$ layers. Consider the data distribution $\mathcal{D}$ and a model $f$ trained on a dataset $\mathrm{D} \subseteq \mathcal{D}^n$ with $n$ samples. Define the generalization gap as
\begin{align}
    &\Delta(f) =\notag\\
    &\E_{\substack{(\mX,Y)\sim\mathcal{D}\\ \mZ_l=f_{1:l}(\mX)\\ \hmY=f_{l+1:L}(\mZ_l)}}\left[\mathcal{L}(Y,\hmY)\right] - \E_{\substack{(\mX,Y)\sim \mathrm{D}\\ \mZ_l=f_{1:l}(\mX)\\ \hmY=f_{l+1:L}(\mZ_l)}}\left[\mathcal{L}(Y,\hmY)\right].
\end{align}
Here, $f_{i:j}$ represents a sub-network that takes input at layer $i$ and produces an output $\mZ_j \in \mathcal{Z}_j$ at layer $j$. We denote $\hmY \in \hat{\mathcal{Y}}$ as the random variable of class likelihoods according to the model, and $\mathcal{L}(Y,\hmY)$ as the loss between the target $Y$ and predictions $\hmY$.

Then, with high probability, the following holds \cite{kawaguchi2023does}:
\begin{align}
    \Delta(f) \in \tilde{O}\left( \sqrt{\frac{I(\mX;\mZ_l\mid Y)+I(f_{1:l};\mathrm{D})}{n}} \right).\label{eq:infobot}
\end{align}

Importantly, the authors show that even when $I(\mX;\mZ_l\mid Y)$ is infinite, such as in the case of some deterministic networks with continuous domains, the generalization bound holds with finite mutual information computed by assuming binning \cite{saxe2019information}. When invoking mutual information in this paper, we assume that binning can be performed to make these quantities finite.

\section{Perturbed Forgetting Perspective of SAM}
\label{sec:perspective}

In this section, we detail our perspective of \emph{perturbed forgetting}, under which we assert that SAM dynamics exhibit forgetting of undesirable model biases through perturbations to benefit generalization.

To start, we consider that the perturbations in SAM seek to exhibit a smaller generalization gap by discarding undesirable biases like spurious relationships. However, the purpose of perturbing is to exhibit a better learning signal in the gradient update. Therefore, the perturbation must not increase the likelihood of the targets, as doing so would dampen the necessary learning signal for the weight update. Due to this constraint, the decrease in generalization gap comes at the expense of increased generalization error, and optimal perturbations schemes should allow attaining a low generalization gap with a minimal increase of error.

\paragraph{Relation to Information Bottleneck.} To understand how SAM perturbations can reduce the generalization gap through forgetting, we utilize the information bottleneck principle and the results of \citet{kawaguchi2023does}. Let us consider the class likelihoods $\hmY$ as the representation in \cref{eq:infobot}. Then, for a model parameterized by $\vtheta$, with high probability the following holds:
\begin{align}
    \Delta(\vtheta) \in \tilde{O}\left( \sqrt{\frac{I(\mX;\hmY\mid Y)+I(\vtheta;\mathrm{D})}{n}} \right).\label{eq:forgetinfobot}
\end{align}
Consider the perturbed parameters from \cref{eq:samperturb} as $\vtheta^p = \argmax_{\norm{\veps}_2 \leq \rho} L_{\vtheta + \veps}(S)$. Denoting the class likelihoods as $\hmY^p$ at $\vtheta^p$, we conjecture that the SAM perturbation reduces both $I(\vtheta^p;\mathrm{D})$ and $I(\mX;\hmY^p\mid Y)$ w.r.t. $\vtheta$.

$I(\vtheta;\mathrm{D})$ quantifies the ability to identify a specific sampling of the training dataset $\mathrm{D}\sim\mathcal{D}$ by observing parameters $\vtheta$. A better fit on the training data allows easier identification of the training dataset from the parameters. In contrast, SAM maximizes the loss and reduces the likelihood of the targets under the model, implying $I(\vtheta^p;\mathrm{D}) \leq I(\vtheta;\mathrm{D})$.

The other term, $I(\mX;\hmY\mid Y)$, quantifies the amount of superfluous information (irrelevant for classification) encoded in the output likelihoods $\hmY$ about the inputs $\mX$. We consider that this superfluous information can provide a view of the biases the model exhibits, and we refer to these biases as \emph{output-exposed biases}. We can write $I(\mX;\hmY\mid Y)$ as
\begin{align}
    I(\mX;\hmY\mid Y) &= \E_{(\vx,\vy) \sim \mathrm{D}} \left[ H\!\left(p_\vtheta(\hmY \mid \vx),p_\vtheta(\hmY \mid y)\right) \right.\notag\\
    &\qquad\qquad\quad\left.- H\!\left(p_\vtheta(\hmY\mid \vx)\right) \right].\label{eq:condmi}
\end{align}
In accordance with the information bottleneck principle, $\hmY=\onehot(Y)$ is one solution to minimizing $I(\mX;\hmY\mid Y)$. However, we need to consider the constraint for learnability of not increasing the likelihood of the target. We argue that SAM can decrease \cref{eq:condmi} under this constraint when the perturbation batch size is small.

\paragraph{Ensembles of Perturbations ($\vm$-SAM).} We interpret $m$-SAM as generating an ensemble of perturbed models per update step using a batch $S$, where the distribution of class likelihoods for an input $\vx \in \mathcal{X}$ is
\begin{align}
    p^\text{$m$-SAM}(\hmY\mid \vx) &= \E_{\veps \sim p_\vtheta(\veps\mid S)}\left[ p_{\vtheta+\veps}(\hmY\mid \vx) \right],
\end{align}
and $\veps$ is a perturbation sampled as
\begin{align}
    \tilde{S}\sim S^m,\quad \veps = \rho \frac{\nabla_\vtheta L_\vtheta(\tilde{S})}{\norm{\nabla_\vtheta L_\vtheta(\tilde{S})}_2}.
\end{align}
However, with a small learning rate, we can also consider an implicit ensemble of perturbations across update steps in full-batch SAM with $\abs{S}=m$. By choosing an appropriate perturbation scheme, a diverse ensemble can minimize \cref{eq:condmi} by increasing the entropy $H(p_\vtheta(\hmY\mid \vx))$ and decreasing the cross-entropy $H(p_\vtheta(\hmY \mid \vx),p_\vtheta(\hmY \mid y))=H(p_\vtheta(\hmY \mid \vx),\E_{\vx'\sim\mathrm{D}(\mX\mid y)}[p_\vtheta(\hmY \mid \vx')])$ by making the high-entropy distributions similar for each input per label.

Note that in addition to diversity, the change in the inductive biases with perturbation influences the quality of the update gradients and their validity at the unperturbed $\vtheta$. Constraining perturbations to an $\epsilon$-neighborhood, as done by SAM, is a simple approach to maintaining this gradient validity.

\paragraph{Small Perturbation Batches (Low $\vm$ in $\vm$-Sharpness).} Unlike the desirable outcome of improved generalization on unseen samples when training to fit larger datasets, the ``generalization'' of maximization with a large perturbation batch beyond its samples can hamper the diversity of perturbations. For instance, requiring maximization to affect a large number of examples simultaneously can limit possible perturbations to those that discard the most prominent globally useful features or simply reduce prediction confidence. On the other hand, perturbing using small batches can offer steepest ascent directions that ``overfit'' differently, introducing the desired noise in the estimation of $\hmY$.

\paragraph{Forgetting Undesirable Biases.} Global maximization is not the goal of perturbed forgetting, as it can generate poor models exhibiting low-quality gradients. Instead, we view maximization on a small number of examples as a mechanism to expose and ``forget'' undesirable shortcuts, or model biases, learned by the model pertaining to those examples. Without this forgetting, the same gradient directions would otherwise contribute to the next update step, potentially causing overfitting. While SAM discards model biases when computing update gradients, it does not immediately unlearn the biases at the unperturbed parameters. We conjecture that SAM will implicitly unlearn them over training by not utilizing them, but we can interpret surrogate gap minimization of GSAM as an explicit mechanism to unlearn forgotten biases. The inconsistency between gradient directions at the original and the perturbed parameters comes from the forgetting of model biases, and minimizing the surrogate gap using \cref{eq:gsamgrad} can be seen as minimizing this inconsistency.

\paragraph{Relation to Other Empirical Observations.} Our perspective of SAM discarding undesirable biases aligns with empirical observations like SAM learning low-rank features \cite{andriushchenko2023sharpness} and reducing harmful overfitting \cite{chen2023does}. As we do not explicitly call for any notion of flatness in the loss surface, our perspective does not clash with the challenges in correlating flatness with generalization \cite{andriushchenko2022towards,andriushchenko2023modern,wen2023sharpness}. Finally, we note that worst-case perturbations, which we view as targeting model biases, have been claimed to be important in SAM \cite{andriushchenko2022towards}. However, in \cref{sec:method}, we design an alternative perturbation to target output-exposed biases, which significantly improves generalization over using steepest ascent perturbations.

\section{Perturb to Forget Output-Exposed Biases}
\label{sec:method}

In this section, we design an alternative perturbation function to forget undesirable model biases in SAM that are exposed through the model's outputs.

\subsection{Setup}

Consider a more general class of extragradient methods \cite{korpelevich1976extragradient,juditsky2011solving,mishchenko2020revisiting} that SAM belongs to, generalizing \cref{eq:samgrad} as:
\begin{align}
\nabla_\vtheta L_\vtheta^{\text{EG}}(S) \approx \nabla_\vtheta L_\vtheta(S) \left|_{\vtheta + \rho \frac{\nabla_\vtheta L^p_\vtheta(S)}{\norm{\nabla_\vtheta L^p_\vtheta(S)}_2}} \right..\label{eq:eggrad}
\end{align}
Here, the perturbed parameters are computed by taking a gradient ascent step to maximize the perturbation objective $L^p_\vtheta(S)$, which equals the task loss $L_\vtheta(S)$ for SAM.

SAM and its variants \cite{zhuang2022surrogate,kwon2021asam,liu2022random,bahri2021sharpness} have commonly been evaluated on tasks such as image classification and language modeling, where the models are trained to maximize the likelihood of discrete outputs such as class predictions or tokens of a sequence. Here, we consider the task of multi-class classification with $C$ classes, where a model parameterized by $\vtheta$ is trained by minimizing the cross-entropy or the sigmoid cross-entropy \cite{beyer2020we} loss between the target label $y \in \{1,\ldots,C\}$ and the model predictions. When using softmax on the model outputs $\vz \in \R^C$ to represent the predicted distribution $\hvy$, the gradient of the cross-entropy loss for a single example can be written as:
\begin{align}
    \nabla_\vtheta \mathcal{L}^\text{CE}(y, \hvy) &= \E_{i\sim \hvy}\left[\nabla_\vtheta z_i\right] - \nabla_\vtheta z_y. \label{eq:ce_grad}
\end{align}

The sigmoid cross-entropy loss, which has been shown to improve ImageNet accuracy, exhibits a similar gradient, but with $\E_{i\sim \hvy}\left[\nabla_\vtheta z_i\right]$ replaced by $\sum_{i=1}^C \hat{y}_i\nabla_\vtheta z_i$ due to $\sum_{i=1}^C\hat{y}_i \neq 1$ in general. However, for ease of notation, we choose to abuse the expectation notation when referring to the gradients of both losses.

\subsection{Output Bias Forgetting (OBF) Perturbation}

Similar to the steepest ascent perturbation discussed in \cref{sec:perspective}, we aim to reduce $I(\vtheta^p;\mathrm{D})$ and $I(\mX;\hmY^p\mid Y)$ at the perturbed parameters $\vtheta^p$ w.r.t. $\vtheta$. We choose to retain decreasing the target likelihood as a way to avoid increasing $I(\vtheta^p;\mathrm{D})$. However, we approach minimizing the superfluous information $I(\mX;\hmY^p\mid Y)$ with considerations to reduce output-exposed biases.

When minimizing the loss by taking a step in the negative direction of \cref{eq:ce_grad}, the non-target logits are chosen based on their current corresponding likelihoods and pushed down. While these semantics are desirable during minimization, maximizing sharpens the non-target predictions to arrive at parameters that potentially amplify the model biases if they are exposed in $\hmY$. Instead, we propose a perturbation function that avoids sharpening the model predictions on maximization, and optionally weakens the exposed model biases when they start being useful in training.

We introduce our output bias forgetting (OBF) perturbation $\mathcal{L}^\text{BF}$, defined for a single example as:
\begin{align}
    \mathcal{L}^\text{BF}(y, \hvy) &= (1-\alpha)\mathcal{L}^\text{CE}(y, \hvy) - \mathcal{L}^\text{CE}(\mathcal{U}, \hvy),\\
    \nabla_\vtheta \mathcal{L}^\text{BF}(y, \hvy) &= \E_{i\sim\mathcal{U}}\left[\nabla_\vtheta z_i\right]\notag\\
    &\,\,\,\,\,- \left(\alpha \E_{i\sim \hvy}\left[\nabla_\vtheta z_i\right] + (1-\alpha) \nabla_\vtheta z_y\right)\!.
\end{align}
Here, $\mathcal{U}$ denotes a uniform distribution and $\alpha \in [0,1]$ controls how much to weaken the model biases. When $\alpha=0$, we avoid explicitly changing the magnitude of the model biases. Such perturbations can be beneficial at the beginning of training, when the model's biases are not useful for the training task but need to be considered to efficiently traverse away from the initialization. When $\alpha=1$, maximizing $\mathcal{L}^\text{BF}$ becomes equivalent to minimizing the cross-entropy loss for a uniform target. Once the model has learned biases that are useful for the training task, but undesirable for generalization, a perturbation towards uniformity can help the model discard these biases in computing the update gradient.

As useful but undesirable model biases could emerge later in training, we propose determining the value of $\alpha$ for each sample during training based on the likelihood $\hat{y}_y$ the model assigns to the ground-truth target $y$:
\begin{align}
    \alpha &= \gamma \max\left(\frac{1-\lambda/\hat{y}_y}{1-\lambda}, 0\right).\label{eq:dynamicalpha}
\end{align}
We treat $\gamma \in [0,1]$ and $\lambda \in [0,1)$ as hyperparameters such that $\alpha$ becomes non-zero if $\hat{y}_y>\lambda$, and increases linearly from $0$ to $\gamma$ as the perplexity $1/\hat{y}_y$ goes from $1/\lambda$ to $1$. A reasonable choice for $\lambda$ is $1/C$ and optimal values of $\gamma$ are either $1$ or close to $0$ depending on the model architecture.

Finally, we note that the complexity of replacing the steepest ascent perturbation with OBF remains the same as standard SAM. The two forward and backward passes dominate the computation time for each iteration. We present the full algorithm utilizing OBF within SAM dynamics in \cref{sec:algorithm}.

\section{Related Work}
\label{sec:related}

We situate our contributions amongst other approaches of explaining the workings of SAM \cite{foret2020sharpness} and other methods of improving generalization by forgetting undesirable model biases.

\paragraph{Understanding SAM.} The original explanation for SAM is based on minimizing the PAC-Bayes upper bound from \citet{foret2020sharpness}. Methods like GSAM \cite{zhuang2022surrogate} and ASAM \cite{kwon2021asam} adapt this bound to propose variants of the SAM algorithm. Often, the importance of minimizing sharpness is assumed, and explanations for the success of SAM comprise of showing how it attains flatter minima \cite{bartlett2023dynamics,wen2022sharpness,ujvary2022rethinking,mollenhoff2023sam,kwon2021asam}. However, the importance of sharpness is debated, as it does not necessarily correlate with generalization error \cite{andriushchenko2023modern,mueller2023normalization,kim2023fantastic} in modern deep neural networks or shallow architectures \cite{wen2023sharpness}. Other factors such as data distribution \cite{wen2023sharpness}, architecture, and hyperparameters play critical roles in success of SAM and its variants \cite{andriushchenko2023modern,wen2023sharpness}. \citet{andriushchenko2022towards} point out that the original PAC-Bayes bound does not explain all the aspects of SAM's success. For example, using the worst-case perturbations instead of average-case as is practically done, only makes this bound less tight. They also suggest that some quantity other than sharpness is implicitly minimized when using small perturbation batches in SAM. Our paper offers a response by highlighting the advantage of smaller perturbation batches from a different perspective. Complementary to our notion of SAM perturbations discarding undesirable model biases to improve generalization in realistic training settings, \citet{chen2023does} formally prove that SAM avoids harmful overfitting in two-layer ReLU convolutional networks. Like us, \citet{baek2024why} identify a different set of principles than sharpness minimization to explain SAM's benefits. They do so in the setting of label noise, attributing SAM's label noise robustness to a dynamic mechanism that learns clean examples before fitting noisy ones.

\paragraph{Forgetting.} ``Forget-and-relearn'' \cite{zhou2022fortuitous} is a general framework that proposes that a mechanism of iteratively forgetting undesirable information and relearning it can improve generalization. This framework encompasses other methods such as iterative magnitude pruning \cite{frankle2018lottery}, knowledge evolution \cite{taha2021knowledge}, and neural iterated learning \cite{ren2020compositional}. Existing forget-and-relearn approaches modify the model parameters in-place, necessitating infrequent forgetting operations and the inefficiency of retraining parts of the network. In contrast, under the perturbed forgetting perspective, the dynamics of SAM allow constructing transient information bottlenecks for computing update gradients without damaging the current learning state at every update step. Like the OBF perturbation, \citet{tiwari2023overcoming} use the gradient of the cross-entropy loss towards a uniform distribution to target model biases to forget. However, they utilize an auxiliary layer for predictions for computing these gradients to avoid affecting the actual model likelihoods. In contrast, we utilize the model likelihoods themselves as the affected likelihoods persist only temporarily per perturbation.

\section{Experiments}
\label{sec:exp}

\subsection{Perturbed Forgetting and Generalization}
\label{sec:forgetvsacc}

\begin{figure*}[t]
\vskip 0.2in
\centering
\includegraphics[width=0.82\textwidth]{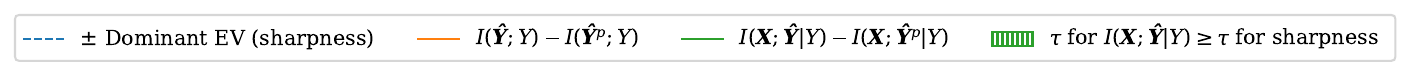}\\
\vspace{-3pt}\begin{subfigure}{0.3\textwidth}
    \centering
    \includegraphics[width=\textwidth]{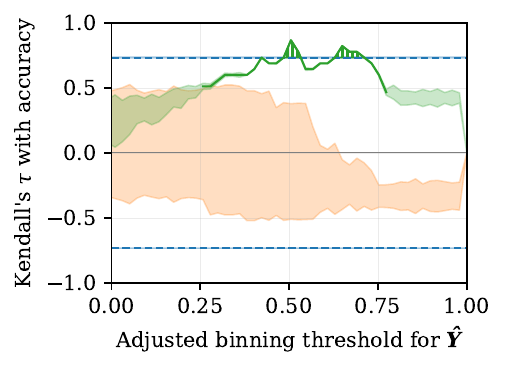}
    \caption{SAM}
\end{subfigure}
\begin{subfigure}{0.3\textwidth}
    \centering
    \includegraphics[width=\textwidth]{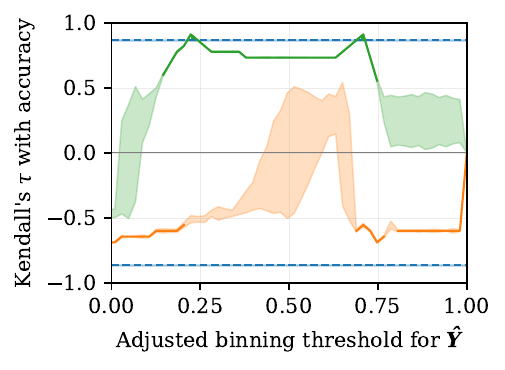}
    \caption{OBF ($\gamma=1$)}
\end{subfigure}
\begin{subfigure}{0.3\textwidth}
    \centering
    \includegraphics[width=\textwidth]{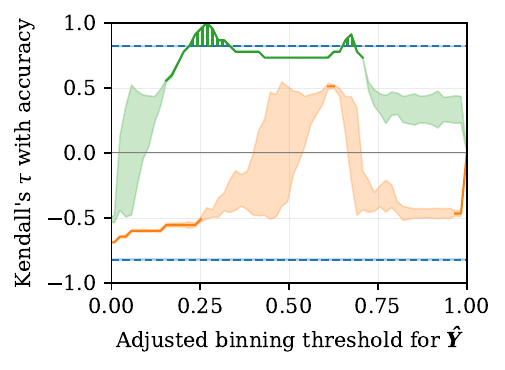}
    \caption{OBF ($\gamma=1/3$)}
\end{subfigure}
\caption{Kendall's $\tau$ correlation of accuracy with sharpness and mutual information metrics averaged over epochs for models trained with different SAM perturbations on CIFAR-10. We train models with perturbation batch size $m\in\{2^k \mid k \in \{0,\ldots,9\}\}$ for each perturbation. Shaded regions indicate the $p$-value estimated with a permutation test, and we show solid lines only when the $p\text{-value} \leq 0.05$.}
\label{fig:mi}
\vskip -0.2in
\end{figure*}

In \cref{sec:perspective,sec:method}, we posited that the superfluous information  quantified by $I(\mX;\hmY\mid Y)$ enables access to the model's biases and perturbing to minimize this quantity with SAM dynamics improves generalization. In this section, we support our claims by measuring the correlation of forgetting output-exposed biases $I(\mX;\hmY\mid Y) - I(\mX;\hmY^p\mid Y)$ with the model's generalization. When $I(\mX;\hmY\mid Y)$ is estimated by thresholding at different values, we find the existence of thresholds that exhibit stronger correlation with generalization than loss surface flatness.

\paragraph{Training.} We train a pool of ViT-S/32 models on CIFAR-10 with three different SAM perturbation strategies: Steepest Ascent (standard SAM), OBF with $\lambda=1/3$, and OBF with $\gamma=0$. For each strategy, we train models with perturbation batch sizes $m\in\{2^k\mid k\in\{0,\ldots,9\}\}$. All models are trained with the same learning rate without decay, and with the same weight decay and perturbation radius $\rho$ hyperparameters to avoid their confounding effects on the training dynamics. We tune these hyperparameters by sweeping across a representative subset of the perturbation settings to ensure that they are comparable to the best-performing hyperparameters for the individual settings. We provide the training hyperparameters in \cref{sec:hyperparams}.

\paragraph{Data Collection.} Unlike sharpness, which can be evaluated on the converged parameters of the model, perturbations are inherently dynamic and need to be captured at various points during training. To this end, we collect the softmax model outputs $\hmY$ on the CIFAR-10 validation set every 25th epoch during training for unperturbed and perturbed parameters for our pool of models.

\paragraph{Mutual Information Estimation.} As the bounds introduced by \citet{kawaguchi2023does} also hold with the assumption of binning, we utilize a simple binning strategy of discretizing the model's softmax outputs to binary based on a threshold. With 10 output dimensions for the 10 classes, the maximum possible number of bins is $2^{10}$. We estimate the mutual information for thresholds $t=10^r$ at 100 values of $r$ linearly spaced between $-12$ to $0$.

For any model checkpoint, the estimated mutual information monotonically increases and then decreases as the binning threshold is increased from $0$ to $1$. Note that as a model is trained, its predictions get sharper and the binning threshold at which the maximum is attained quickly becomes much smaller than chance. We focus on the higher-magnitude variations in model outputs to understand the impact of forgetting, which are captured at thresholds greater than the one exhibiting the highest mutual information.

Different checkpoints, including those for different epochs of the same training run, attain the maximum at different thresholds. To allow comparison across epochs and different values of $m$, we first normalize the maximum to $1$. Then, we adjust the binning thresholds by resampling such that the unperturbed mutual information decreases linearly from $1$ to $0$ between thresholds $0$ to $1$. Consequently, any specific adjusted binning threshold exhibits the same unperturbed $I(\mX;\hmY\mid Y)$ across all collected checkpoints.

\paragraph{Correlating Forgetting with Generalization.} We average the difference $I(\mX;\hmY\mid Y) - I(\mX;\hmY^p\mid Y)$ across all adjusted thresholds for every model and epoch per perturbation type. We evaluate the Kendall rank correlation between this difference and the final CIFAR-10 test accuracy the model attains. We follow the same procedure to also evaluate $I(\hmY; Y) - I(\hmY^p; Y)$, which quantifies the change in task-relevant information when perturbing. Finally, we calculate the correlation between sharpness and the same accuracy for comparison, where sharpness is the Hessian's dominant eigenvalue estimated using power iteration.

\paragraph{Results.} We present the estimated correlations for each adjusted threshold value and perturbation type in \cref{fig:mi}. Discretizing at the thresholds with the highest correlation of accuracy with $I(\mX;\hmY\mid Y) - I(\mX;\hmY^p\mid Y)$ (green curves) reveals undesirable information encoded in the model outputs, that if targeted for forgetting, leads to improved generalization. At these thresholds, we find the correlation of accuracy with forgetting information about the classification target (orange curves) to remain negative, further indicating that the generalization benefits come from discarding superfluous information. Finally, we highlight the regions where forgetting correlates with generalization (green curves) more strongly than flatness (blue lines) with a green hatch pattern.

Our results demonstrate the existence of output-exposed biases and the generalization benefit of forgetting them.

\subsection{Standard Benchmarks}
\label{sec:stdbench}

\begin{table*}[t]
\caption{Top-$1$ accuracies on ImageNet and robustness datasets. For SAM and GSAM, models are trained with standard steepest ascent (\textsc{Steep}) and output bias forgetting (\textsc{OBF}) perturbations. Sharpness (dominant eigenvalue) is estimated for each model using power iteration. Standard deviations are reported with three trials.}
\label{tab:imagenet}
\vskip 0.15in
\begin{center}
\begin{small}
\begin{sc}
\begin{tabular}{@{}lllcccccc@{}}
\toprule
\multirow{2}[2]{*}{Model} &\multirow{2}[2]{*}{Method} &\multirow{2}[2]{*}{Perturb} &\multicolumn{5}{c}{ImageNet-} &\multirow{2}[2]{*}{Sharpness} \\
\cmidrule{4-8}
& & &V1 &Real &V2 &R &Sketch \\
\midrule
\multirow{5}[3]{*}{ViT-S/32} &AdamW &None &69.29{\scriptsize$\pm$0.26} &75.31{\scriptsize$\pm$0.28} &55.48{\scriptsize$\pm$0.58} &19.02{\scriptsize$\pm$0.47} &16.38{\scriptsize$\pm$0.34} &165.6{\scriptsize$\pm$15.2} \\
\cmidrule{2-3}
&\multirow{2}{*}{SAM} &Steep &72.77{\scriptsize$\pm$0.06} &78.89{\scriptsize$\pm$0.05} &58.81{\scriptsize$\pm$0.33} &21.63{\scriptsize$\pm$0.23} &19.68{\scriptsize$\pm$0.50} &14.9{\scriptsize$\pm$1.1} \\
& &OBF &\textbf{74.49{\scriptsize$\pm$0.04}} &{81.31{\scriptsize$\pm$0.05}} &\textbf{61.13{\scriptsize$\pm$0.18}} &\textbf{25.31{\scriptsize$\pm$0.41}} &\textbf{22.58{\scriptsize$\pm$0.13}} &3.9{\scriptsize$\pm$1.4} \\
\cmidrule{2-3}
&\multirow{2}{*}{GSAM} &Steep &73.41{\scriptsize$\pm$0.05} &79.48{\scriptsize$\pm$0.08} &59.94{\scriptsize$\pm$0.15} &22.18{\scriptsize$\pm$0.15} &20.28{\scriptsize$\pm$0.15} &11.6{\scriptsize$\pm$1.2} \\
& &OBF &74.41{\scriptsize$\pm$0.12} &\textbf{81.41{\scriptsize$\pm$0.11}} &\textbf{61.08{\scriptsize$\pm$0.18}} &{25.15{\scriptsize$\pm$0.23}} &22.24{\scriptsize$\pm$0.07} &\textbf{3.1{\scriptsize$\pm$0.7}} \\
\midrule
\multirow{5}[3]{*}{ViT-S/16} &AdamW &None &74.30{\scriptsize$\pm$0.10} &80.04{\scriptsize$\pm$0.04} &61.28{\scriptsize$\pm$0.09} &20.25{\scriptsize$\pm$0.27} &18.15{\scriptsize$\pm$0.11} &59.0{\scriptsize$\pm$9.4} \\
\cmidrule{2-3}
&\multirow{2}{*}{SAM} &Steep &78.73{\scriptsize$\pm$0.08} &85.47{\scriptsize$\pm$0.05} &66.98{\scriptsize$\pm$0.14} &25.69{\scriptsize$\pm$0.09} &24.10{\scriptsize$\pm$0.33} &\textbf{2.4{\scriptsize$\pm$0.4}} \\
& &OBF &\textbf{80.30{\scriptsize$\pm$0.13}} &86.14{\scriptsize$\pm$0.13} &68.62{\scriptsize$\pm$0.05} &27.19{\scriptsize$\pm$0.27} &\textbf{26.45{\scriptsize$\pm$0.22}} &17.1{\scriptsize$\pm$2.8} \\
\cmidrule{2-3}
&\multirow{2}{*}{GSAM} &Steep &78.95{\scriptsize$\pm$0.13} &84.31{\scriptsize$\pm$0.06} &66.80{\scriptsize$\pm$0.43} &24.92{\scriptsize$\pm$0.43} &24.41{\scriptsize$\pm$0.52} &6.0{\scriptsize$\pm$0.3} \\
& &OBF &\textbf{80.32{\scriptsize$\pm$0.06}} &\textbf{86.26{\scriptsize$\pm$0.07}} &\textbf{68.83{\scriptsize$\pm$0.12}} &\textbf{27.48{\scriptsize$\pm$0.10}} &25.92{\scriptsize$\pm$0.13} &4.3{\scriptsize$\pm$1.4} \\
\midrule
\multirow{5}[3]{*}{ResNet-50} &SGD &None &76.86{\scriptsize$\pm$0.07} &83.28{\scriptsize$\pm$0.11} &65.00{\scriptsize$\pm$0.14} &20.29{\scriptsize$\pm$0.36} &20.53{\scriptsize$\pm$0.46} &230.4{\scriptsize$\pm$42.7} \\
\cmidrule{2-3}
&\multirow{2}{*}{SAM} &Steep &77.49{\scriptsize$\pm$0.06} &83.78{\scriptsize$\pm$0.05} &65.26{\scriptsize$\pm$0.21} &21.08{\scriptsize$\pm$0.16} &21.18{\scriptsize$\pm$0.32} &{170.1{\scriptsize$\pm$18.9}} \\
& &OBF &\textbf{77.67{\scriptsize$\pm$0.07}} &84.01{\scriptsize$\pm$0.03} &65.70{\scriptsize$\pm$0.45} &{21.63{\scriptsize$\pm$0.18}} &\textbf{22.17{\scriptsize$\pm$0.26}} &\textbf{164.4{\scriptsize$\pm$25.0}} \\
\cmidrule{2-3}
&\multirow{2}{*}{GSAM} &Steep &77.43{\scriptsize$\pm$0.12} &83.79{\scriptsize$\pm$0.19} &65.37{\scriptsize$\pm$0.26} &21.37{\scriptsize$\pm$0.21} &21.52{\scriptsize$\pm$0.56} &171.0{\scriptsize$\pm$16.8} \\
& &OBF &\textbf{77.66{\scriptsize$\pm$0.08}} &\textbf{84.09{\scriptsize$\pm$0.07}} &\textbf{66.01{\scriptsize$\pm$0.09}} &\textbf{21.76{\scriptsize$\pm$0.23}} &\textbf{22.26{\scriptsize$\pm$0.47}} &\textbf{161.4{\scriptsize$\pm$10.9}} \\
\midrule
\multirow{5}[3]{*}{ResNet-101} &SGD &None &78.44{\scriptsize$\pm$0.08} &84.39{\scriptsize$\pm$0.02} &66.61{\scriptsize$\pm$0.19} &22.91{\scriptsize$\pm$0.83} &23.45{\scriptsize$\pm$1.31} &228.1{\scriptsize$\pm$29.8} \\
\cmidrule{2-3}
&\multirow{2}{*}{SAM} &Steep &79.09{\scriptsize$\pm$0.08} &85.05{\scriptsize$\pm$0.09} &67.24{\scriptsize$\pm$0.20} &23.64{\scriptsize$\pm$0.38} &24.80{\scriptsize$\pm$0.20} &\textbf{155.1{\scriptsize$\pm$12.0}} \\
& &OBF &79.27{\scriptsize$\pm$0.06} &85.17{\scriptsize$\pm$0.10} &{67.85{\scriptsize$\pm$0.17}} &24.21{\scriptsize$\pm$0.26} &\textbf{25.56{\scriptsize$\pm$0.47}} &170.4{\scriptsize$\pm$2.1} \\
\cmidrule{2-3}
&\multirow{2}{*}{GSAM} &Steep &79.11{\scriptsize$\pm$0.04} &85.00{\scriptsize$\pm$0.08} &67.52{\scriptsize$\pm$0.21} &23.65{\scriptsize$\pm$0.39} &24.79{\scriptsize$\pm$0.13} &{166.1{\scriptsize$\pm$2.8}} \\
& &OBF &\textbf{79.40{\scriptsize$\pm$0.07}} &\textbf{85.37{\scriptsize$\pm$0.16}} &\textbf{68.05{\scriptsize$\pm$0.35}} &\textbf{24.52{\scriptsize$\pm$0.10}} &\textbf{25.44{\scriptsize$\pm$0.33}} &{165.7{\scriptsize$\pm$28.4}} \\
\bottomrule
\end{tabular}
\end{sc}
\end{small}
\end{center}
\vskip -0.1in
\end{table*}

We now study the generalization benefits of OBF by comparing models trained with varying architectures and perturbation schemes on standard benchmarks. We also present additional baselines and settings, including ASAM \cite{kwon2021asam}, in \cref{sec:moreexps}.

\paragraph{Datasets.} We train our models on ImageNet-1K, also known as ImageNet-V1 \cite{deng2009imagenet}, and also perform finetuning experiments with CIFAR-10 and CIFAR-100 \cite{krizhevsky2009learning}. When training from scratch, we evaluate on the ImageNet validation set, and the additional test sets ImageNet-Real \cite{beyer2020we} and ImageNet-V2 \cite{recht2019imagenet}. ImageNet-Real corrects idiosyncrasies and errors in the labeling of the original validation set and ImageNet-V2 contains newly-collected data following the original ImageNet data creation process. Additionally, we evaluate our models on the out-of-distribution robustness benchmarks ImageNet-R \cite{hendrycks2021many}, which contains renditions of 200 ImageNet classes in various forms, and ImageNet-Sketch \cite{wang2019learning}, which has black-and-white sketch images for every ImageNet class. For our transfer learning experiments, we evaluate the models on the test splits of CIFAR-\{10,100\}.

\paragraph{Models.} We run our experiments with two model families: vision transformers (ViT) \cite{dosovitskiy2020image} and residual networks (ResNet) \cite{he2016deep}. For ViT, we experiment with ViT-S/32 and ViT-S/16, and choose ResNet-50 and ResNet-101 for the ResNet experiments.

\paragraph{Training.} We follow the setting of GSAM \cite{zhuang2022surrogate} and \citet{chen2021vision}, and train our models with Inception-style pre-processing \cite{szegedy2015going} without strong data augmentations for both ViT and ResNet models. All models are trained with a global batch size of 4096, perturbation batch size $m=64$, and linear learning rate decay schedule with warmup. We apply the the same scheduling of the perturbation radius $\rho$ that GSAM uses for both GSAM and SAM, which provide stronger baseline results, but keep $\rho$ constant when using the OBF perturbation. We provide all hyperparameter values in \cref{sec:hyperparams}.

\paragraph{Finetuning.} When finetuning on CIFAR-\{10,100\}, we use the same pre-processing scheme as we do for training. We finetune ViT-S/32 and ResNet-50 with SGD with momentum 0.9 for 100 epochs, without weight decay, and gradients clipped to global norm 1. We use a smaller batch size of 512, but keep the perturbation batch size $m=64$. All other hyperparameters are provided in \cref{sec:hyperparams}.

\paragraph{Metrics.} We report the generalization performance as the classification top-1 accuracy on the selected evaluation datasets. Additionally, we also report sharpness of our reported models, which is the Hessian's dominant eigenvalue estimated using the power iteration method. We report standard deviations for our metrics where available with three trials when training from scratch and six trials when finetuning. The six finetuning trials comprise of three groups of two finetuning trials, each group finetuning a model from one of the three training trials.

\subsubsection{ImageNet Generalization}

We present our results on ImageNet evaluation and robustness benchmarks in \cref{tab:imagenet}. First, we note that training SAM with the OBF perturbation generalizes better than using standard steepest ascent perturbations with either SAM or GSAM in a majority of the studied benchmarks and methods. Additionally, utilizing GSAM with OBF further improves results in most settings for ViTs. Under the perturbed forgetting perspective, both steepest ascent perturbations and OBF target model biases for forgetting, but the kinds of biases exposed through the probing mechanisms they utilize can be different. Our results suggest that the outputs of ViTs provide significantly better access to its undesirable biases compared to its steepest ascent directions, whereas the improvements of using one perturbation over another is small for ResNets.

Furthermore, our results also indicate that the models that converge to the flattest regions of the loss surface seldom perform the best. Moreover, despite OBF outperforming steepest ascent perturbations in most settings, it only exhibits lowest sharpness in the case of ViT-S/32. Our results support the notion that the training dynamics of SAM are critical for generalization. We also remark that while SAM and GSAM work best with tricks like scheduling the perturbation radius, the OBF perturbation outperforms them without resorting to doing so.

\subsubsection{Transfer Learning to CIFAR Datasets}

\begin{table}[t]
\caption{Transfer learning top-$1$ accuracies of models finetuned on CIFAR-\{10,100\} after pretraining on ImageNet, where either pretraining (\textsc{Pre}) or finetuning (\textsc{FT}) uses SAM. Standard deviations are reported with six trials.}
\label{tab:cifar}
\vskip 0.15in
\begin{center}
\begin{small}
\begin{sc}
\begin{tabular}{@{}lllcc@{}}\toprule
\multirow{2}[2]{*}{Model} &\multicolumn{2}{c}{SAM Perturb} &\multicolumn{2}{c}{CIFAR-} \\\cmidrule{2-5}
&Pre &FT &10 &100 \\\midrule
\multirow{4}[2]{*}{\makecell{ViT-S\\/32}} &\multirow{2}{*}{None} &Steep &97.74{\scriptsize$\pm$0.08} &86.94{\scriptsize$\pm$0.09} \\
& &OBF &\textbf{97.79{\scriptsize$\pm$0.04}} &\textbf{87.04{\scriptsize$\pm$0.13}} \\
\cmidrule{2-5}
&Steep &\multirow{2}{*}{None} &97.69{\scriptsize$\pm$0.04} &86.21{\scriptsize$\pm$0.17} \\
&OBF & &\textbf{97.92{\scriptsize$\pm$0.05}} &\textbf{86.99{\scriptsize$\pm$0.10}} \\
\midrule
\multirow{4}[2]{*}{\makecell{ResNet\\-50}} &\multirow{2}{*}{None} &Steep &96.84{\scriptsize$\pm$0.12} &83.29{\scriptsize$\pm$0.25} \\
& &OBF &\textbf{96.91{\scriptsize$\pm$0.10}} &\textbf{83.41{\scriptsize$\pm$0.23}} \\
\cmidrule{2-5}
&Steep &\multirow{2}{*}{None} &96.16{\scriptsize$\pm$0.16} &81.81{\scriptsize$\pm$0.21} \\
&OBF & &\textbf{96.40{\scriptsize$\pm$0.24}} &\textbf{81.91{\scriptsize$\pm$0.23}} \\
\bottomrule
\end{tabular}
\end{sc}
\end{small}
\end{center}
\vskip -0.1in
\end{table}

We study the transfer learning capability of models trained on ImageNet with SAM with our studied perturbations, as well as the ability to finetune ImageNet-pretrained models with these methods. We present our transfer learning results for CIFAR-10 and CIFAR-100 in \cref{tab:cifar}.

We observe that models pretrained with OBF allow improved transfer to CIFAR-10 and CIFAR-100. The advantages of OBF can also be seen when finetuning models pretrained without SAM in all settings but one.

\subsubsection{Effect of $\gamma$ and $\lambda$}
\label{sec:effecthyper}

\begin{figure}[t]
\vskip 0.2in
\begin{center}
\centerline{
\includegraphics[width=0.49\linewidth]{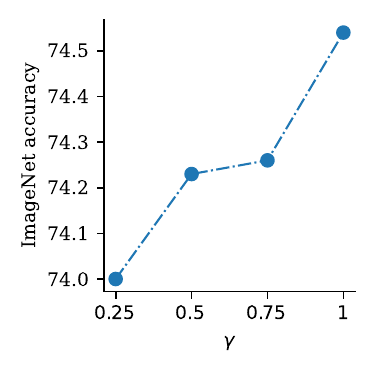}
\includegraphics[width=0.49\linewidth]{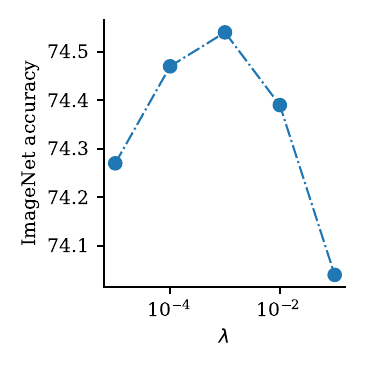}
}
\caption{Effect of the hyperparameters $\gamma$ (with fixed $\lambda=1/C$, where $C$ is the number of classes) and $\lambda$ (with fixed $\gamma=1$) on ImageNet top-$1$ accuracy for ViT-S/32 trained using the output bias forgetting (OBF) perturbation in SAM.}
\label{fig:hyperparams}
\end{center}
\vskip -0.2in
\end{figure}

The hyperparameters $\gamma$ and $\lambda$ control the conditions for activating a push towards a uniform model prediction during OBF. We find that their optimal values differ substantially by model architecture. For ResNets, we find the best value of $\gamma$ to be close to zero, making it insensitive to the choice of $\lambda$. This setting weighs the gradients for each non-target prediction equally, unlike steepest ascent perturbations which weigh the gradients proportionally to the predicted likelihoods. For ViTs, we find it useful to allow an explicit pressure towards uniform predictions by setting $\gamma=1$. Both settings avoid sharpening the predictions when perturbing, as sharpening can exacerbate output-exposed biases.

\cref{fig:hyperparams} shows how ImageNet performance changes with varying $\gamma$ and $\lambda$ with ViT-S/32. When $\gamma$ is low, the forgetting strength is lower, and the performance gain from perturbed forgetting is lower. With $\gamma=1$, $\lambda$ determines the target likelihood threshold for each sample beyond which the perturbation pushes predictions towards uniformity. When $\lambda$ is too high, the push towards uniformity does not start until much later in training, missing opportunities to improve generalization. On the other hand, when $\lambda$ is too small, the push starts before the model learns useful mappings, hurting training efficiency and, consequently, generalization. The optimal value $1/C$, where $C$ is the number of classes, suggests that such explicit forgetting is beneficial when it targets model biases that become available once the model starts performing better than chance for a given sample.

\section{Discussion}
\label{sec:discussion}

Generalization to the true data distribution can be understood as generalization across plausible structural variations of the data. Accordingly, the flatness we desire is in the space of these variations, not necessarily in the parameter space. There is no inherent causal link between these two notions of flatness without additional assumptions. Relating our contributions to sharpness minimization, we argue that perturbed forgetting can provide a framework for pursuing flatness in the space of data variations by reducing sensitivity to sampling of the training dataset. Small perturbation batches can reveal shortcuts the model has learned for the included examples, and gathering gradients after forgetting encourages the model to instead rely on and strengthen a global structure that benefits a larger set of examples.

Our paper uses the perturbed forgetting perspective to devise a perturbation that can outperform standard SAM, GSAM, and ASAM (in \cref{sec:moreexps}). However, we do not claim OBF to be an optimal perturbation for all settings. Both steepest ascent and OBF perturbations allow perturbed forgetting of model biases, albeit potentially of different biases. Furthermore, like many prior works on SAM \cite{foret2020sharpness,kwon2021asam,liu2022random}, our paper is limited to image classification experiments. Undesirable model biases can be easier to target and forget with different perturbations under different architectures and training domains. Characterizing the exact nature of the biases targeted by different perturbations and a more formal theoretical treatment of perturbed forgetting can provide insights for generalizing in a wide range of settings, including those where standard SAM is ineffective or difficult to integrate.

\section*{Acknowledgements}
We are grateful to the anonymous reviewers for their valuable feedback during the review period.

\section*{Impact Statement}
This paper presents work whose goal is to advance the field of machine learning. There are many potential societal consequences of our work, none which we feel must be specifically highlighted here.

\bibliography{main}
\bibliographystyle{icml2024}

\newpage
\appendix
\counterwithin{figure}{section}
\counterwithin{table}{section}
\counterwithin{equation}{section}
\counterwithin{algorithm}{section}
\onecolumn

\section{Optimization Algorithm}
\label{sec:algorithm}

{%
\centering%
\vspace{-6pt}
\begin{minipage}{0.6\linewidth}%
\begin{algorithm}[H]
   \caption{Iterated Output Bias Forgetting}
   \label{alg:forgetting}
\begin{algorithmic}
   \STATE {\bfseries Input:} Training dataset $\mathrm{D} = \cup_{i=1}^{\abs{\mathrm{D}}}\{\vx^{(i)},y^{(i)}\} \subseteq \mathcal{D} \subseteq \mathcal{X} \times \{1,\ldots,C\}$, likelihood function $f_\vtheta$, perturbation step size $\rho$, update step size $\eta$, bias forgetting strength hyperparameters $\gamma$ and $\lambda$.
   \STATE Initialize weights $\vtheta_0$, step $t=0$;
   \REPEAT
        \STATE Sample $(\vx,y)$ from $\mathrm{D}$;
        \STATE $\hvy = f_{\vtheta_t}(\vx)$;\hfill\COMMENT{Current likelihood}
        \STATE $\alpha = \gamma \max\left(\frac{1-\lambda/\hat{y}_y}{1-\lambda}, 0\right)$;\hfill\COMMENT{Forgetting strength}
        \STATE $\Delta_t=\nabla_{\vtheta_t} \left[(1-\alpha)\mathcal{L}^\text{CE}(y, \hvy) - \mathcal{L}^\text{CE}(\mathcal{U}, \hvy)\right]$;
        \STATE $\vtheta^p_t = \vtheta_t + \rho \frac{\Delta_t}{\norm{\Delta_t}}$;\hfill\COMMENT{Perturbed weights}
        \STATE $\vtheta_{t+1}=\vtheta_t - \eta \nabla_{\vtheta^p_t} \mathcal{L}^\text{CE}\left(y,f_{\vtheta^p_t}(\vx)\right)$;\hfill\COMMENT{Update}
    \UNTIL{convergence}
\end{algorithmic}
\end{algorithm}
\end{minipage}%
\par%
\vspace{8pt}%
}

We present the optimization algorithm incorporating OBF within SAM dynamics, without perturbation ensembling for simplicity, in \cref{alg:forgetting}.

\section{Additional Experiments}
\label{sec:moreexps}

\subsection{Standard Benchmarks}

We provide additional results on ViT-S/32 and ResNet-50 in \cref{tab:moreimagenet}, with the following new settings compared to \cref{tab:imagenet}.

\paragraph{ASAM.} We compare OBF with the standard steepest ascent perturbations in the ASAM \cite{kwon2021asam} framework. ASAM with OBF outperforms standard ASAM on all benchmarks, and yet exhibits higher sharpness. Our results further support the importance of the forgetting mechanism over loss surface flatness of the parameters.

\paragraph{Shrink and Perturb.} Ideas related to forgetting have also been explored in settings of non-stationarity, with a goal of restoring plasticity under continually changing data distributions \cite{ash2020warm,elsayed2023utility,dohare2023maintaining}. Although non-stationarity is not a problem in our evaluation settings and SAM does not perform in-place forgetting of the parameters, we consider a shrink-and-perturb \cite{ash2020warm} baseline for empirical comparison. To ease hyperparameter search and maintain reasonable parameter value scales, we adopt modifications from \citet{doro2023sampleefficient,kumar2023maintaining} to perform shrinking and perturbing simultaneously through the linear interpolation of parameters towards their initialization. We perturb the parameters for every update using the tuned interpolation factor of $10^{-5}$. We find shrink-and-perturb to remain comparable with the vanilla baselines, suggesting a generalization advantage of perturbed forgetting over in-place forgetting.

\paragraph{Random SAM Perturbations.} We add a SAM baseline that samples perturbations uniformly on a unit hypersphere of radius $\rho$. Without an explicit mechanism to target undesirable model biases, we find that SAM with random perturbations exhibits performance closer to vanilla training than SAM with steepest ascent or OBF perturbations.

\paragraph{Label Smoothing.} Finally, we consider a vanilla baseline with label smoothing of 0.1, which provides an alternative approach for encouraging uniform predictions to improve generalization. We find label smoothing to improve performance over the vanilla settings, but it does not outperform our SAM settings. We note that label smoothing deviates from our other settings by changing the training objective, adding constraints not otherwise imposed in fitting the training data.

\begin{table}[t]
\caption{Top-$1$ accuracies on ImageNet and robustness datasets. Without SAM, models are trained in either a vanilla setting, with per-step shrink-and-perturb (\textsc{ShrinkPerturb}), or with label smoothing of $0.1$ (\textsc{LabelSmooth}). SAM and ASAM models are trained with standard steepest ascent (\textsc{Steep}) and output bias forgetting (\textsc{OBF}) perturbations. SAM is also trained with random (\textsc{Random}) perturbations. Sharpness (dominant eigenvalue) is estimated for each model using power iteration. Standard deviations are reported with three trials. $^\dagger$\emph{Note:} Label smoothing produces a different training objective and loss surface geometry compared to the other settings.}
\label{tab:moreimagenet}
\begin{center}
\begin{small}
\begin{sc}
\resizebox{\linewidth}{!}{%
\begin{tabular}{@{}lllcccccc@{}}
\toprule
\multirow{2}[2]{*}{Model} &\multirow{2}[2]{*}{Method} &\multirow{2}[2]{*}{Perturb} &\multicolumn{5}{c}{ImageNet-} &\multirow{2}[2]{*}{Sharpness} \\
\cmidrule{4-8}
& & &V1 &Real &V2 &R &Sketch \\
\midrule
\multirow{8}[3]{*}{\makecell{ViT-S\\/32}} &AdamW &\multirow{3}{*}{None} &69.29{\scriptsize$\pm$0.26} &75.31{\scriptsize$\pm$0.28} &55.48{\scriptsize$\pm$0.58} &19.02{\scriptsize$\pm$0.47} &16.38{\scriptsize$\pm$0.34} &165.6{\scriptsize$\pm$15.2} \\
&ShrinkPerturb & &69.05{\scriptsize$\pm$0.07} &75.29{\scriptsize$\pm$0.08} &55.45{\scriptsize$\pm$0.32} &18.99{\scriptsize$\pm$0.11} &16.14{\scriptsize$\pm$0.23} &325.8{\scriptsize$\pm$290.9} \\
&LabelSmooth$^\dagger$ & &69.75{\scriptsize$\pm$0.10} &75.94{\scriptsize$\pm$0.10} &55.95{\scriptsize$\pm$0.30} &19.66{\scriptsize$\pm$0.16} &16.82{\scriptsize$\pm$0.20} &1959.9{\scriptsize$\pm$1319.8} \\
\cmidrule{2-3}
&\multirow{3}{*}{SAM} &Steep &72.77{\scriptsize$\pm$0.06} &78.89{\scriptsize$\pm$0.05} &58.81{\scriptsize$\pm$0.33} &21.63{\scriptsize$\pm$0.23} &19.68{\scriptsize$\pm$0.50} &14.9{\scriptsize$\pm$1.1} \\
& &OBF &{74.49{\scriptsize$\pm$0.04}} &\textbf{81.31{\scriptsize$\pm$0.05}} &\textbf{61.13{\scriptsize$\pm$0.18}} &\textbf{25.31{\scriptsize$\pm$0.41}} &\textbf{22.58{\scriptsize$\pm$0.13}} &\textbf{3.9{\scriptsize$\pm$1.4}} \\
& &Random &69.23{\scriptsize$\pm$0.28} &75.43{\scriptsize$\pm$0.31} &55.27{\scriptsize$\pm$0.26} &19.03{\scriptsize$\pm$0.27} &16.45{\scriptsize$\pm$0.37} &147.4{\scriptsize$\pm$30.4} \\
\cmidrule{2-3}
&\multirow{2}{*}{ASAM} &Steep &74.45{\scriptsize$\pm$0.11} &81.23{\scriptsize$\pm$0.11} &60.78{\scriptsize$\pm$0.25} &24.07{\scriptsize$\pm$0.12} &21.68{\scriptsize$\pm$0.23} &6.5{\scriptsize$\pm$0.4} \\
& &OBF &\textbf{74.73{\scriptsize$\pm$0.19}} &{81.24{\scriptsize$\pm$0.25}} &60.95{\scriptsize$\pm$0.28} &24.65{\scriptsize$\pm$0.26} &22.40{\scriptsize$\pm$0.10} &30.3{\scriptsize$\pm$11.6} \\
\midrule
\multirow{8}[3]{*}{\makecell{ResNet\\-50}} &SGD &\multirow{3}{*}{None} &76.86{\scriptsize$\pm$0.07} &83.28{\scriptsize$\pm$0.11} &65.00{\scriptsize$\pm$0.14} &20.29{\scriptsize$\pm$0.36} &20.53{\scriptsize$\pm$0.46} &230.4{\scriptsize$\pm$42.7} \\
&ShrinkPerturb & &76.83{\scriptsize$\pm$0.03} &83.28{\scriptsize$\pm$0.10} &64.62{\scriptsize$\pm$0.27} &20.25{\scriptsize$\pm$0.31} &20.97{\scriptsize$\pm$0.37} &256.6{\scriptsize$\pm$31.7} \\
&LabelSmooth$^\dagger$ & &77.18{\scriptsize$\pm$0.31} &{83.93{\scriptsize$\pm$0.21}} &{65.53{\scriptsize$\pm$0.19}} &21.25{\scriptsize$\pm$0.30} &21.11{\scriptsize$\pm$0.01} &277.8{\scriptsize$\pm$6.3} \\
\cmidrule{2-3}
&\multirow{3}{*}{SAM} &Steep &77.49{\scriptsize$\pm$0.06} &83.78{\scriptsize$\pm$0.05} &65.26{\scriptsize$\pm$0.21} &21.08{\scriptsize$\pm$0.16} &21.18{\scriptsize$\pm$0.32} &{170.1{\scriptsize$\pm$18.9}} \\
& &OBF &{77.67{\scriptsize$\pm$0.07}} &{84.01{\scriptsize$\pm$0.03}} &{65.70{\scriptsize$\pm$0.45}} &{21.63{\scriptsize$\pm$0.18}} &{22.17{\scriptsize$\pm$0.26}} &{164.4{\scriptsize$\pm$25.0}} \\
& &Random &77.00{\scriptsize$\pm$0.10} &83.27{\scriptsize$\pm$0.11} &64.76{\scriptsize$\pm$0.15} &20.56{\scriptsize$\pm$0.26} &20.94{\scriptsize$\pm$0.24} &220.4{\scriptsize$\pm$10.9} \\
\cmidrule{2-3}
&\multirow{2}{*}{ASAM} &Steep &77.30{\scriptsize$\pm$0.02} &84.07{\scriptsize$\pm$0.03} &65.55{\scriptsize$\pm$0.16} &21.71{\scriptsize$\pm$0.02} &21.75{\scriptsize$\pm$0.15} &\textbf{33.6{\scriptsize$\pm$2.99}} \\
& &OBF &\textbf{78.17{\scriptsize$\pm$0.07}} &\textbf{84.66{\scriptsize$\pm$0.05}} &\textbf{66.55{\scriptsize$\pm$0.15}} &\textbf{23.84{\scriptsize$\pm$0.12}} &\textbf{24.21{\scriptsize$\pm$0.42}} &39.1{\scriptsize$\pm$1.28} \\
\bottomrule
\end{tabular}}
\end{sc}
\end{small}
\end{center}
\vskip -0.1in
\end{table}

\subsection{Ablating Dynamic vs. Fixed OBF $\alpha$}

\begin{table}[t]
\caption{Ablating output bias forgetting (OBF) hyperparameters for SAM with ViT-S/32.}
\label{tab:fixedalpha}
\vskip 0.15in
\begin{center}
\begin{small}
\begin{sc}
\begin{tabular}{@{}ccc@{}}
\toprule
SAM Perturb &OBF Hyperparams &ImageNet-V1 \\
\midrule
Steep & {\scriptsize N/A} &72.81 \\
\cmidrule{1-2}
\multirow{8}[2]{*}{OBF} & $\alpha=0$ &73.33 \\
& $\alpha=10^{-5}$ &72.73 \\
& $\alpha=10^{-4}$ &72.49 \\
& $\alpha=10^{-3}$ &73.11 \\
& $\alpha=10^{-2}$ &73.65 \\
& $\alpha=10^{-1}$ &73.77 \\
& $\alpha=1$ &74.15 \\
\cmidrule{2-2}
& $\gamma=1$, $\lambda=10^{-3}$ &\textbf{74.53} \\
\bottomrule
\end{tabular}
\end{sc}
\end{small}
\end{center}
\vskip -0.1in
\end{table}

The OBF hyperparameters $\lambda$ and $\gamma$ are used to dynamically produce a value of $\alpha$ per sample based on \cref{eq:dynamicalpha}. In this section, we perform an ablation experiment with fixed values of $\alpha$ with ViT-S/32 models trained on ImageNet. Our results in \cref{tab:fixedalpha} confirm that determining $\alpha$ dynamically leads to the best performance in our setting. Additionally, we also see fixed values of $\alpha$ that exhibit better generalization than steepest ascent perturbations with SAM.

\section{Training Details and Hyperparameters}
\label{sec:hyperparams}

\begin{table}[tp]
\caption{Hyperparameters used for training models for each task, optimization algorithm, and perturbation type.}
\label{tab:hyperparams}
\vskip 0.15in
\begin{center}
\begin{small}
\begin{sc}
\resizebox{\linewidth}{!}{%
\begingroup
\setlength{\tabcolsep}{2.75pt}
\begin{tabular}{@{}ccccccccccccccccccccc@{}}\toprule
\normalsize\multirow{2}[5]{*}{Model} &\normalsize\multirow{2}[5]{*}{Task} &\normalsize\multirow{2}[5]{*}{\makecell{Batch\\Size}} &\multicolumn{2}{c}{\normalsize Epochs} &\normalsize \multirow{2}[5]{*}{Optim} &\normalsize \multirow{2}[5]{*}{Loss} &\normalsize \multirow{2}[5]{*}{\makecell{Weight\\Decay}} &\multicolumn{2}{c}{\normalsize\makecell{Learning\\Rate}} &\normalsize \multirow{2}[5]{*}{\makecell{Clip\\Grad}} &\normalsize \multirow{2}[5]{*}{\makecell{Head\\Bias\\Init}} &\normalsize \multirow{2}[5]{*}{Algo} &\normalsize \multirow{2}[5]{*}{Perturb} &\multicolumn{2}{c}{\normalsize $\rho$} &\multicolumn{2}{c}{\normalsize OBF} &\multicolumn{2}{c}{\normalsize GSAM} &\normalsize\multirow{2}[5]{*}{\makecell{ASAM\\Fixed\\Norm}} \\\cmidrule{4-5}\cmidrule{9-10}\cmidrule{15-20}
&\normalsize  &\normalsize  &\normalsize Train &\normalsize Warm &\normalsize  &\normalsize  &\normalsize  &\normalsize Max &\normalsize Min &\normalsize  &\normalsize  &\normalsize  &\normalsize  &\normalsize Max &\normalsize Min &\normalsize $\lambda$ &\normalsize $\gamma$ &\normalsize $\alpha$ &\normalsize \makecell{Norm\\Backup} \\\midrule
\multirow{13}[13]{*}{\makecell{ViT-S\\/32}} &\multirow{3}[2]{*}{\makecell{CIFAR\\Forget\\vs. Acc}} &\multirow{3}[2]{*}{512} &\multirow{3}[2]{*}{300} &\multirow{3}[2]{*}{32} &\multirow{3}[2]{*}{AdamW} &\multirow{3}[2]{*}{BCE} &\multirow{3}[2]{*}{1.2} &\multicolumn{2}{c}{\multirow{3}[2]{*}{$3 \times 10^{-4}$}} &\multirow{3}[2]{*}{1.0} &\multirow{3}[2]{*}{$-10$} &\multirow{3}[2]{*}{SAM} &Steep &\multicolumn{2}{c}{\multirow{3}[2]{*}{0.2}} &\multicolumn{2}{c}{{\scriptsize N/A}} &\multicolumn{2}{c}{\multirow{3}[2]{*}{{\scriptsize N/A}}} &\multirow{3}[2]{*}{{\scriptsize N/A}} \\\cmidrule{14-14}\cmidrule{17-18}
& & & & & & & & & & & & &\multirow{2}{*}{OBF} & & &\multirow{2}{*}{$1/3$} &0 & & \\
& & & & & & & & & & & & & & & & &1 & & \\\cmidrule{2-21}
&\multirow{7}[7]{*}{\makecell{Image-\\Net\\Train}} &\multirow{7}[7]{*}{4096} &\multirow{7}[7]{*}{300} &\multirow{7}[7]{*}{32} &\multirow{7}[7]{*}{AdamW} &\multirow{7}[7]{*}{BCE} &\multirow{7}[7]{*}{0.3} &\multirow{7}[7]{*}{0.003} &\multirow{7}[7]{*}{$3 \times 10^{-5}$} &\multirow{7}[7]{*}{1.0} &\multirow{7}[7]{*}{$-10$} &\multicolumn{2}{c}{None} &\multicolumn{2}{c}{{\scriptsize N/A}} &\multicolumn{2}{c}{\multirow{2}[2]{*}{{\scriptsize N/A}}} &\multicolumn{2}{c}{\multirow{3}[4]{*}{{\scriptsize N/A}}} &\multirow{5}[5]{*}{{\scriptsize N/A}} \\\cmidrule{13-16}
& & & & & & & & & & & &\multirow{2}[2]{*}{SAM} &Steep &0.6 &0 & & & & \\\cmidrule{17-18}
& & & & & & & & & & & & &OBF &\multicolumn{2}{c}{0.6} &$1/C$ &1 & & \\\cmidrule{13-20}
& & & & & & & & & & & &\multirow{2}{*}{GSAM} &Steep &0.6 &0 &\multicolumn{2}{c}{{\scriptsize N/A}} &\multirow{2}{*}{0.4} &No \\
& & & & & & & & & & & & &OBF &\multicolumn{2}{c}{0.6} &$1/C$ &1 & &Yes \\\cmidrule{13-21}
& & & & & & & & & & & &\multirow{2}{*}{ASAM} &Steep &\multicolumn{2}{c}{\multirow{2}{*}{6.0}} &\multicolumn{2}{c}{{\scriptsize N/A}} &\multicolumn{2}{c}{\multirow{2}{*}{{\scriptsize N/A}}} &\multirow{2}{*}{No} \\
& & & & & & & & & & & & &OBF & & &$1/C$ &$10^{-12}$ & & & \\\cmidrule{2-21}
&\multirow{3}[3]{*}{\makecell{CIFAR\\Fine-\\tune}} &\multirow{3}[3]{*}{512} &\multirow{3}[3]{*}{100} &\multirow{3}[3]{*}{5} &\multirow{3}[3]{*}{SGD} &\multirow{3}[3]{*}{BCE} &\multirow{3}[3]{*}{0} &\multirow{3}[3]{*}{0.003} &\multirow{3}[3]{*}{0} &\multirow{3}[3]{*}{1.0} &\multirow{3}[3]{*}{0} &\multicolumn{2}{c}{None} &\multicolumn{2}{c}{{\scriptsize N/A}} &\multicolumn{2}{c}{\multirow{2}[2]{*}{{\scriptsize N/A}}} &\multicolumn{2}{c}{\multirow{3}[3]{*}{{\scriptsize N/A}}} &\multirow{3}[3]{*}{{\scriptsize N/A}} \\\cmidrule{13-16}
& & & & & & & & & & & &\multirow{2}[2]{*}{SAM} &Steep &\multicolumn{2}{c}{\multirow{2}[2]{*}{0.05}} & & & & \\\cmidrule{17-18}
& & & & & & & & & & & & &OBF & & &$1/C$ &1 & & \\\midrule
\multirow{5}[5]{*}{\makecell{ViT-S\\/16}} &\multirow{5}[5]{*}{\makecell{Image-\\Net\\Train}} &\multirow{5}[5]{*}{4096} &\multirow{5}[5]{*}{300} &\multirow{5}[5]{*}{32} &\multirow{5}[5]{*}{AdamW} &\multirow{5}[5]{*}{BCE} &\multirow{5}[5]{*}{0.3} &\multirow{5}[5]{*}{0.003} &\multirow{5}[5]{*}{$3 \times 10^{-5}$} &\multirow{5}[5]{*}{1.0} &\multirow{5}[5]{*}{$-10$} &\multicolumn{2}{c}{None} &\multicolumn{2}{c}{{\scriptsize N/A}} &\multicolumn{2}{c}{\multirow{2}[2]{*}{{\scriptsize N/A}}} &\multicolumn{2}{c}{\multirow{3}[3]{*}{{\scriptsize N/A}}} &\multirow{5}[5]{*}{{\scriptsize N/A}} \\\cmidrule{13-16}
& & & & & & & & & & & &\multirow{2}[2]{*}{SAM} &Steep &\multicolumn{2}{c}{\multirow{2}[2]{*}{0.6}} & & & & \\\cmidrule{17-18}
& & & & & & & & & & & & &OBF & & &$1/C$ &1 & & \\\cmidrule{13-20}
& & & & & & & & & & & &\multirow{2}{*}{GSAM} &Steep &0.6 &0 &\multicolumn{2}{c}{{\scriptsize N/A}} &1 &No \\
& & & & & & & & & & & & &OBF &\multicolumn{2}{c}{0.6} &$1/C$ &1 &0.4 &Yes \\\midrule
\multirow{10}[10]{*}{\makecell{ResNet\\-50}} &\multirow{7}[7]{*}{\makecell{Image-\\Net\\Train}} &\multirow{7}[7]{*}{4096} &\multirow{7}[7]{*}{90} &\multirow{7}[7]{*}{16} &\multirow{7}[7]{*}{SGD} &\multirow{7}[7]{*}{CE} &\multirow{7}[7]{*}{0.001} &\multirow{7}[7]{*}{1.6} &\multirow{7}[7]{*}{0.016} &\multirow{7}[7]{*}{{\scriptsize N/A}} &\multirow{7}[7]{*}{0} &\multicolumn{2}{c}{None} &\multicolumn{2}{c}{{\scriptsize N/A}} &\multicolumn{2}{c}{\multirow{2}[2]{*}{{\scriptsize N/A}}} &\multicolumn{2}{c}{\multirow{3}[3]{*}{{\scriptsize N/A}}} &\multirow{5}[5]{*}{{\scriptsize N/A}} \\\cmidrule{13-16}
& & & & & & & & & & & &\multirow{2}[2]{*}{SAM} &Steep &0.04 &0.02 & & & & \\\cmidrule{17-18}
& & & & & & & & & & & & &OBF &\multicolumn{2}{c}{0.04} &$1/C$ &$10^{-12}$ & & \\\cmidrule{13-20}
& & & & & & & & & & & &\multirow{2}{*}{GSAM} &Steep &0.04 &0.02 &\multicolumn{2}{c}{{\scriptsize N/A}} &\multirow{2}{*}{0.01} &Yes \\
& & & & & & & & & & & & &OBF &\multicolumn{2}{c}{0.04} &$1/C$ &$10^{-12}$ & &No \\\cmidrule{13-21}
& & & & & & & & & & & &\multirow{2}{*}{ASAM} &Steep &0.8 &0.4 &\multicolumn{2}{c}{{\scriptsize N/A}} &\multicolumn{2}{c}{\multirow{2}{*}{{\scriptsize N/A}}} &\multirow{2}{*}{Yes} \\
& & & & & & & & & & & & &OBF &\multicolumn{2}{c}{0.8} &$1/C$ &$10^{-12}$ & & & \\\cmidrule{2-21}
&\multirow{3}[3]{*}{\makecell{CIFAR\\Fine-\\tune}} &\multirow{3}[3]{*}{512} &\multirow{3}[3]{*}{100} &\multirow{3}[3]{*}{5} &\multirow{3}[3]{*}{SGD} &\multirow{3}[3]{*}{CE} &\multirow{3}[3]{*}{0} &\multirow{3}[3]{*}{0.01} &\multirow{3}[3]{*}{0} &\multirow{3}[3]{*}{1.0} &\multirow{3}[3]{*}{0} &\multicolumn{2}{c}{None} &\multicolumn{2}{c}{{\scriptsize N/A}} &\multicolumn{2}{c}{\multirow{2}[2]{*}{{\scriptsize N/A}}} &\multicolumn{2}{c}{\multirow{3}[3]{*}{{\scriptsize N/A}}} &\multirow{3}[3]{*}{{\scriptsize N/A}} \\\cmidrule{13-16}
& & & & & & & & & & & &\multirow{2}[2]{*}{SAM} &Steep &\multicolumn{2}{c}{\multirow{2}[2]{*}{0.1}} & & & & \\\cmidrule{17-18}
& & & & & & & & & & & & &OBF & & &$1/C$ &$10^{-12}$ & & \\\midrule
\multirow{5}[5]{*}{\makecell{ResNet\\-101}} &\multirow{5}[5]{*}{\makecell{Image-\\Net\\Train}} &\multirow{5}[5]{*}{4096} &\multirow{5}[5]{*}{90} &\multirow{5}[5]{*}{16} &\multirow{5}[5]{*}{SGD} &\multirow{5}[5]{*}{CE} &\multirow{5}[5]{*}{0.001} &\multirow{5}[5]{*}{1.6} &\multirow{5}[5]{*}{0.016} &\multirow{5}[5]{*}{{\scriptsize N/A}} &\multirow{5}[5]{*}{0} &\multicolumn{2}{c}{None} &\multicolumn{2}{c}{{\scriptsize N/A}} &\multicolumn{2}{c}{\multirow{2}[2]{*}{{\scriptsize N/A}}} &\multicolumn{2}{c}{\multirow{3}[3]{*}{{\scriptsize N/A}}} &\multirow{5}[5]{*}{{\scriptsize N/A}} \\\cmidrule{13-16}
& & & & & & & & & & & &\multirow{2}[2]{*}{SAM} &Steep &0.04 &0.02 & & & & \\\cmidrule{17-18}
& & & & & & & & & & & & &OBF &\multicolumn{2}{c}{0.04} &$1/C$ &$10^{-12}$ & & \\\cmidrule{13-20}
& & & & & & & & & & & &\multirow{2}{*}{GSAM} &Steep &0.04 &0.02 &\multicolumn{2}{c}{{\scriptsize N/A}} &\multirow{2}{*}{0.01} &Yes \\
& & & & & & & & & & & & &OBF &\multicolumn{2}{c}{0.04} &$1/C$ &$10^{-12}$ & &No \\
\bottomrule
\end{tabular}
\endgroup%
}
\end{sc}
\end{small}
\end{center}
\vskip -0.1in
\end{table}

We provide the hyperparameters used in our experiments in \cref{tab:hyperparams}, with descriptions and details of its columns below:
\begin{itemize}
    \item \smallsc{Model}.\, The model architecture.
    \item \smallsc{Task}.\, The task the hyperparameters are for. \smallsc{CIFAR Forget vs. Acc} provides settings for training the pool of models for \cref{sec:forgetvsacc}, and \smallsc{ImageNet Train} and \smallsc{CIFAR Finetune} provide the training and finetuning hyperparameters for \cref{sec:stdbench}.
    \item \smallsc{Batch Size}.\, The global batch size used for computing the update step. This is separate from the perturbation batch size $m$, which is set to $64$ in all cases except in \cref{sec:forgetvsacc}.
    \item \smallsc{Epochs}.\, The total training epochs are provided under \smallsc{Train}, and \smallsc{Warm} provides the number of warmup epochs for a linear learning rate decay schedule with linear warmup.
    \item \smallsc{Optim}.\, The base optimizer used. Here, \smallsc{SGD} uses momentum 0.9. Both \smallsc{AdamW} and \smallsc{SGD} use decoupled weight decay \cite{loshchilov2017decoupled}.
    \item \smallsc{Loss}.\, The loss function used for computing the update gradients. \smallsc{CE} is the cross-entropy loss and \smallsc{BCE} is the sigmoid cross-entropy \cite{beyer2020we} loss.
    \item \smallsc{Weight Decay}.\, The weight decay strength.
    \item \smallsc{Learning Rate}.\, The maximum (\smallsc{Max}) and minimum (\smallsc{Min}) learning rates for the linear learning rate schedule.
    \item \smallsc{Clip Grad}.\, Gradients are clipped to norm of this value before taking the update step. Gradient clipping is disabled if this value is \smallsc{N/A}.
    \item \smallsc{Head Bias Init}.\, The initial value for the bias parameters of the classification head.
    \item \smallsc{Algo}.\, The SAM-like algorithm for which these hyperparameters are for. A value of \smallsc{None} indicates vanilla training.
    \item \smallsc{Perturb}.\, The perturbation type, which can be steepest ascent (\smallsc{Steep}), output bias forgetting (\smallsc{OBF}), or \smallsc{None} in case of vanilla training.
    \item \smallsc{$\rho$}.\, The perturbation neighborhood size for SAM-like algorithms. When \smallsc{Max} and \smallsc{Min} have different values, $\rho$ is decayed linearly with a linear warmup using the same scheme as the learning rate.
    \item \smallsc{OBF}.\, The hyperparameters $\lambda$ and $\gamma$ for the \smallsc{OBF} perturbation. Here, $C$ in the $\lambda$ values indicates the number of classes, which are 1000, 100, and 10 for ImageNet, CIFAR-100, and CIFAR-10 respectively.
    \item \smallsc{GSAM}.\, The hyperparameters for GSAM \cite{zhuang2022surrogate}. We were unable to reproduce the officially reported numbers using the authors' hyperparameters. Although we report lower performance for GSAM with ViT, we outperform or match the authors in all other settings including the vanilla and SAM baselines. Additionally, we achieve improved performance with GSAM in some settings by normalizing the perturbing gradient before decomposing it. This normalization is performed if \smallsc{Norm Backup} is \smallsc{Yes}.
    \item \smallsc{ASAM Fixed Norm}.\, When using ASAM \cite{kwon2021asam} with ResNet-50, we outperform the authors' reported numbers by ensuring that the perturbation always has a fixed norm by applying the inverse normalization operator on the gradients before normalizing, and not after. For ViT, we achieve the best baseline performance by applying it after normalization, and not before.
\end{itemize}

\end{document}